# Automated Floodwater Depth Estimation Using Large Multimodal Model for Rapid Flood Mapping

Temitope Akinboyewa, Huan Ning, M. Naser Lessani, Zhenlong Li[*]

Geoinformation and Big Data Research Laboratory

Department of Geography, The Pennsylvania State University, University Park, PA, 16801

[*]zhenlong@psu.edu

**Abstract:** Information on the depth of floodwater is crucial for rapid mapping of areas affected by floods. However, previous approaches for estimating floodwater depth, including field surveys, remote sensing, and machine learning techniques, can be time-consuming and resource-intensive. This paper presents an automated and fast approach for estimating floodwater depth from on-site flood photos. A pre-trained large multimodal model, GPT-4 Vision, was used specifically for estimating floodwater. The input data were flooding photos that contained referenced objects, such as street signs, cars, people, and buildings. Using the heights of the common objects as references, the model returned the floodwater depth as the output. Results show that the proposed approach can rapidly provide a consistent and reliable estimation of floodwater depth from flood photos. Such rapid estimation is transformative in flood inundation mapping and assessing the severity of the flood in near-real time, which is essential for effective flood response strategies.

## 1 Introduction

Flooding is one of the most common and devastating natural hazards, leading to significant human and economic losses annually (Bentivoglio et al., 2022). As climate change contributes to more frequent and intense precipitation events, flooding severity is expected to increase (Tabari, 2020). In the face of such a threat, rapid estimation of flooded areas becomes crucial. During flooding events, emergency managers need timely and accurate information about inundated areas to coordinate response operations effectively (Manjusree et al., 2012; Li et al., 2020)

Rapid flood mapping not only provides an immediate understanding of the extent and severity of flooding but also aids authorities and humanitarian organizations in allocating and distributing essential resources like food, water, and medical supplies (Cohen et al., 2018). Identifying flooded areas quickly is also important for protecting critical infrastructure such as power plants and water treatment facilities, thereby minimizing disruption and speeding up recovery efforts (J. Li et al., 2021). Additionally, flood maps are essential in analyzing flood patterns and informing urban planning and flood mitigation strategies (Meghanadh et al., 2020).

Floodwater depth is an important factor in flood inundation mapping (Fohringer, et al., 2015, Z. Li et al., 2018). Information about the depth of floodwater is also crucial for assessing the severity of floods and evaluating flood risk mitigation measures. This information is essential for deploying rescue efforts, determining road closures, and assessing accessible areas (Cohen et al., 2018). Furthermore, it also has a key role in supporting emergency response, assessing accessibility and designing suitable intervention plans, calculating water volumes, allocating resources for water pumping, and rapidly estimating the costs for intervention and reconstruction (Cian et al., 2018). Data on flood depths is not only useful for immediate response but also for post-disaster analysis, including evaluating property damage and assessing flood risks (Nguyen et al., 2016).

Various techniques have been used to estimate flood depth. Conventional approaches such as field surveys have been utilized to determine flood depth by directly measuring high-water marks in affected areas (Chaudhary et al., 2020). Though this method has a very high precision, it is time-consuming and labor-intensive. Additionally, it is limited to small-scale applications and can be impacted by weather conditions (Elkhrachy, 2022). Conventional methods also rely on information from stream gauges at specific locations to offer real-time flood data, such as water level. However, these approaches have constraints when the floodwater exceeds the height of the gauge and when scattered gauge placements are unable to adequately cover flooded areas (Z. Li et al., 2018). Another approach is to use hydrodynamic models to assess floodwater depth, including the Hydrologic Engineering Center's River Analysis System (HEC-RAS) (Athira et al., 2023; Brunner, 2016), Delft-3D (Haq et al., 2020), and LISFLOOD-FP (Yin et al., 2022). These models are known for their accuracy in simulating complex flood dynamics (Elkhrachy, 2022). However, their utilization is hindered by the requirement for extensive input datasets that involve detailed topographic, meteorological, and hydrological data. Additionally, these models require significant computational resources and rely on powerful computing systems.

Remote sensing data has been used extensively for flood management in recent decades. Large-scale knowledge about the extent of floods can be obtained through remote sensing data, such as satellite imagery. To determine floodwater depth, studies have combined optical and Synthetic Aperture Radar (SAR) with a Digital Elevation Model (DEM). For example, Cian et al (2018) introduced a semi-automatic approach to calculate flood depth, by utilizing SAR imagery and statistical estimation of DEM from LIDAR (Light Detection and Ranging). Surampudi and Kumar (2023) utilized SAR data and Shuttle Radar Topography Mission (SRTM) DEM to generate water depth that closely follows surface undulations in agricultural lands. These methods appear to be efficient in estimating floodwater depth. However, image acquisition is limited by the temporal resolution of satellites (Bovenga et al., 2018). Also, optical sensors can be affected by cloud cover during flood events (Chaudhary et al., 2020), making it impossible to have flood images for areas covered with clouds. Additionally, vertical inaccuracies are frequently produced by DEMs, especially over complicated terrain such as urban areas. As a result, they are unreliable in identifying the important topographical elements which determine how floods behave (Schumann, 2014).

Recent advancements in machine learning have significantly revolutionized flood depth estimation. Numerous studies have employed sophisticated computer vision algorithms to remotely assess water levels. For instance, Pan et al. (2018) employed a Convolutional Neural Network (CNN)-based methodology to monitor the length of a ruler in footage captured by a video camera strategically placed adjacent to a river. Similarly, utilizing a mask region-based convolution neural network (Mask R-CNN), Park et al. (2021) achieved flood depth estimation by detecting submerged vehicles in flood photos. Furthermore, the wealth of flood images on social media platforms has provided a rich source for researchers to employ computer vision algorithms in estimating floodwater depth. Feng et al. (2020) introduced a workflow focusing on retrieving images containing humans from social media to estimate water levels. In a distinct approach, Quan et al. (2020) matched water levels with human poses, categorizing flood severity into "above the knee" and "below the knee". Other innovative approaches involve the integration of deep learning techniques with web images for flood depth estimation (Meng et al., 2019), leveraging CNNs to segment stop signs and extract flood depth data from images featuring such signs (Song and Tuo, 2021), and utilizing an object detection model based on CNN for automatically estimating water depth from images sourced from social media (Li et al., 2023). While machine learning models have showcased their effectiveness in flood depth estimation, it is crucial to acknowledge their dependency on substantial, annotated training datasets. Creating such datasets can be a resource-intensive and time-consuming endeavor, underscoring a notable challenge in the practical implementation of these models.

The recent advent of generative artificial intelligence (AI) models, notably the Generative Pre-trained Transformer (GPT), has emerged as an exceptional development. These models exhibit a remarkable capability to comprehend human natural language, enabling proficient task execution across diverse domains. GPT-4 Vision (GPT-4 hereafter), a large-scale multimodal model, has demonstrated several impressive abilities of vision-language understanding and generation (OpenAI, 2023). For example, GPT-4 can generate natural language descriptions of images and even perform image processing tasks from descriptions written as text. These models can also provide intelligent solutions that are more similar to human thinking, enabling us to use general artificial intelligence to solve problems in various applications (Wen et al., 2023).

In geographic information science, researchers have explored the potential of GPT with applications to image generation, captioning, and analysis assistance in visuals, to name a few (Osco et al., 2023). A notable effort by Hu et al. (2023) involves the integration of geo-knowledge with GPT models for identifying location descriptions and their respective categories. This fusion results in a geo-knowledge-guided GPT model good at accurately extracting location descriptions from disaster-related social media messages. Li and Ning (2023) pioneered a prototype of autonomous GIS (Geographic Information System) utilizing the GPT-4 API, aiming to accept tasks through natural language and autonomously solve spatial problems. Other endeavors include exploring the potential of GPT-4 in map-making (Tao and Xu, 2023) and examining the capability of GPT-4 in extracting information from streetview photographs in unban analytics (Crooks and Chen, 2024).

Given the exceptional capabilities of GPT-4, its anticipated impact extends to various fields, including flood risk management. This research presents an automated, fast, and reliable approach leveraging GPT-4 to estimate floodwater depth from photographs capturing flood events. This study aims to make a substantive contribution to disaster management and emergency response, with the potential to enhance mitigation strategies, ultimately contributing to life-saving efforts and minimizing economic losses.

## 2 Method

### 2.1 Overview of the proposed approach

The GPT-4 model, developed by OpenAI, was trained using increasingly large amounts of data and has proven to be highly effective at extracting valuable information from images, even without requiring a separate training dataset. In this study, we propose a novel, fully automated framework for estimating the floodwater depth by leveraging the advanced potential of GPT-4. This framework, FloodDepth-GPT, uses a GPT-4 model Python API to estimate the floodwater depth. The overall concept of the proposed approach is illustrated in Figure 1. The approach begins by inputting flooding photos containing objects that can serve as consistent indicators for reference. Such street objects can include vehicles, humans, and street signs. By assessing the known height and relative submersion of these objects, FloodDepth-GPT can estimate water levels according to the visible objects within the photos. For instance, if the water reaches the knee level of a person whose height is known, FloodDepth-GPT can "deduce" the depth of water based on this comparative analysis. Besides the water depth, the FloodDepth-GPT also outlines the rationale behind it, which enhances the transparency, understanding, and explainability of the process.

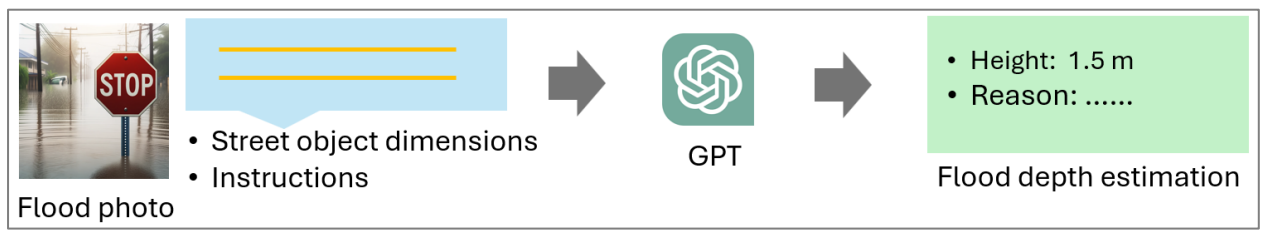

**Figure 1. Workflow of FloodDepth-GPT**

### 2.2 Design of FloodDepth-GPT

The FloodDepth-GPT is a customized GPT with a set of prompts structured to guide the tool specifically toward estimating floodwater depth. These prompts include directions related to identifying and measuring reference points in the image, assessing visible waterlines on objects, and identifying known heights of common objects, such as humans, vehicles, and stop signs, present in the image (Appendix 1). The standard heights of the various objects were specified. For example, the average height of a man and the different parts of the body (i.e., knee, waist, shoulder, height, and waist) were included in the prompt (see Tables 1, 2, and 3 for the heights of different reference objects). Figure 2 shows samples of the objects used in this study with their corresponding heights.

**Table 1: Average height of different parts of a human body (Fryar et al., 2021; J. Li et al., 2023)**

| Part of the body | Description | Male (m) | Female (m) |
|---|---|---|---|
| Knee height | Distance from the bottom of the feet to the top of the knee | 0.4 | 0.4 |
| Weist height | Distance from the ground to a person's waist | 0.9 | 0.8 |
| Shoulder height | Distance from the ground to the top of a person's shoulder | 1.4 | 1.3 |
| Head | Total height (distance in meters from the ground to the top of the head | 1.75 | 1.60 |

**Table 2: Average height of different parts of a vehicle (Neighbor Storage, 2023)**

| Vehicle heights | Description | Sedan (m) | Truck (m) | Bus (m) |
|---|---|---|---|---|
| Ground clearance | The height of the base of the car from the ground | 0.2 | 0.5 | 0.7 |
| Ride height | The height of the lower side of the car's body (just above the wheel) from the ground | 0.6 | 0.8 | 1.0 |
| Windshield's Height | The height at the start of the windshield from the ground | 1.0 | 1.3 | 2.0 |
| Total height | The total height of the car | 1.4 | 1.8 | 3.2 |

**Table 3: Stop sign dimensions (Song and Tuo, 2021; U.S. Department of Transportation, 2023)**

| Part of the stop sign | Description | Height (m) |
|---|---|---|
| Stop sign | The height of the stop sign itself, from the top edge to the bottom edge. | 0.9 |
| Pole height | The height of the bottom of the sign from the ground | 2.0 |
| Total height | The total height of the car | 2.9 |

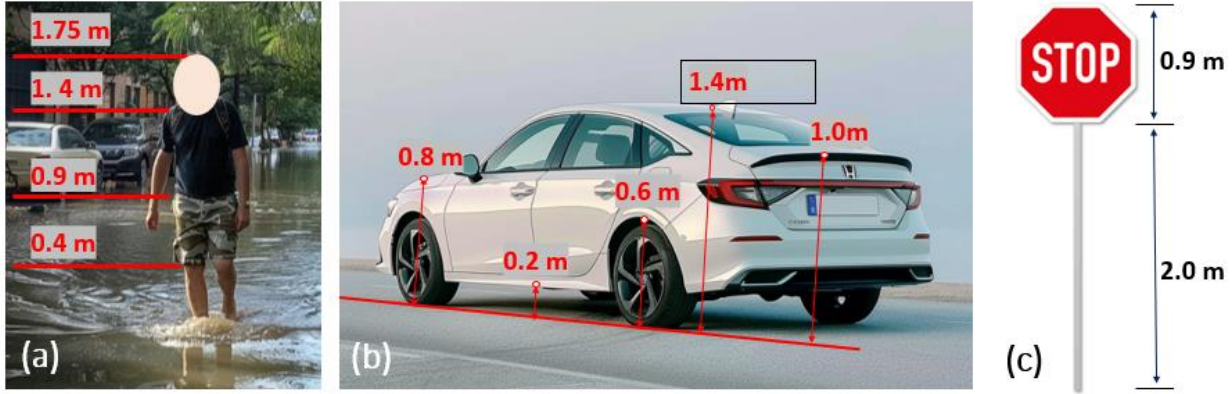

**Figure 2. Sample of reference objects with heights**

Another crucial output of FloodDepth-GPT were detailed explanations of its estimations. This involves clear communication of the visual cues used in the estimation process and presentation of the depth measurements for ease of understanding and global applicability, enhancing the explainability of the AI output. Finally, the model was instructed to avoid speculation and base its analyses on the available objects within the image.

## 2.3  Performance evaluation

The ability of the FloodDepth-GPT was examined as follows. We collected 150 flood photos from various online sources. Previous studies have utilized flood photos to estimate the depth of floodwater based on different reference objects, including stop signs, vehicles, and humans (J. Li et al., 2023). This experimental dataset also incorporates these three components as main reference objects (Tables 1, 2, and 3). We ensured that each selected photo has at least one of these objects. These photos served as input into FloodDepth-GPT, and the floodwater depth estimation for each photo was obtained using the model.

To evaluate the performance of the GPT model, this study compared the floodwater depth estimated by FloodDepth-GPT (GPT Estimation) and floodwater depth estimated manually by three individuals (Manual Estimation). The manual estimation processes were conducted independently, and used the average heights detailed in Tables 1-3. Furthermore, the Mean Absolute Error (MAE) was calculated to quantitatively measure the accuracy of the FloodDepth-GPT estimations in comparison to the manual estimations. The MAE was computed using the formula provided in Equation 1, where $m_i$ is the manual-estimated depth, $gpt_i$ is the FloodDepth-GPT estimation, and $n$ is the number of images.

$$MAE = \sum_{i=1}^{n} \frac{m_i - gpt_i}{n} \qquad (1)$$

## 3  Results

Figure 3 presents the correlation between the flood water depth estimation of GPT and human. Result shows that there is a strong positive correlation between GPT estimation and the average estimations from human observers (Pearson's correlation coefficient $r = 0.8894$). Additionally, there is a strong correlation between GPT and each human estimation ($r = 0.8705, 0.8585, 0.8742$ for human 1, 2, and 3, respectively). Overall, the data points are clustered along the regression line, which suggests that the estimations made by GPT are consistent with the estimations given by humans. Also, the consistency across the human estimations lends credibility to their use as a benchmark for evaluating the accuracy of GPT.

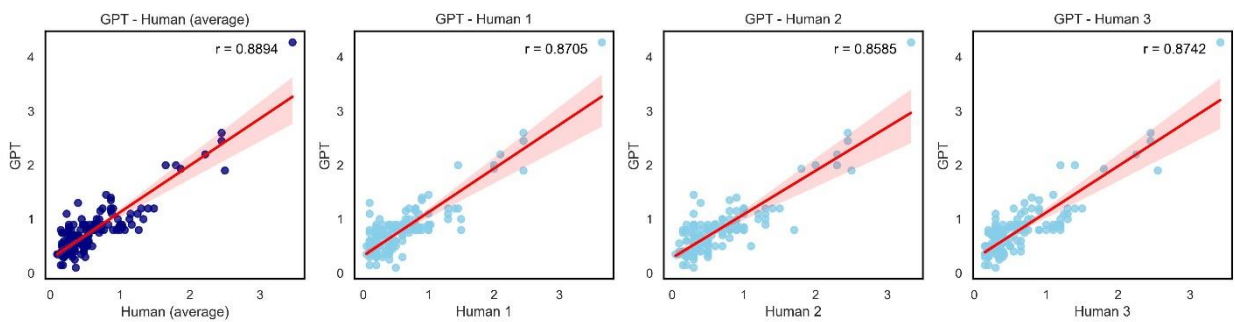

**Figure 3. Scatter plots of flood depth estimation of GPT and human (unit: meter).**

Additionally, comparing the GPT-based estimations and the manual estimations, a Mean Absolute Error (MAE) of 27 cm was recorded. This is consistent with the reported MAE - ranging from 7cm to 31cm - in previous studies (Alizadeh Kharazi and Behzadan, 2021; Chaudhary et al., 2020; Park et al., 2021), which uses deep learning methods in estimating floodwater depth from images obtained from social media platforms.

Samples of FloodDepth-GPT estimations and responses are presented in Figure 4. These results are in line with manual estimations. Furthermore, Figure 5 highlights a detailed sample response from the FloodDepth-GPT model, showing its ability not only to provide reliable floodwater depth estimations but also to give reasonable explanations behind the estimations. This process entails identifying reference objects and the water level in flood photos, then utilizing the known height of these identified objects to estimate the depth of the floodwater. FloodDepth-GPT, for example, successfully identified the truck in a flood photo and estimated that the water level of the flood was below the bottom of the truck's door. Utilizing the identified height of the truck's bottom level, FloodDepth-GPT accurately estimated the floodwater depth.

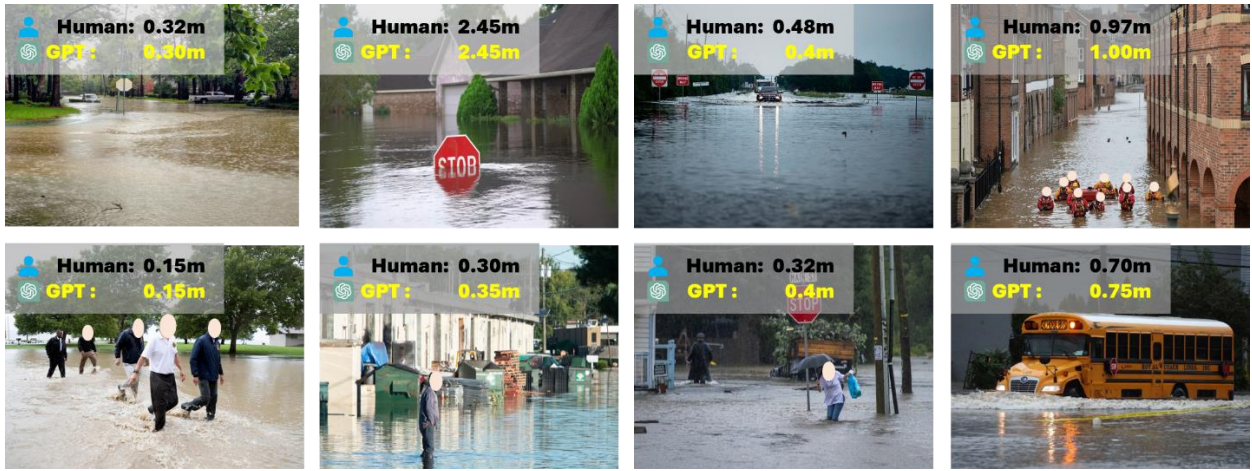

**Figure 4. Selected flood depth estimation results by human and GPT**

To estimate the floodwater depth in the provided image, let's use the pickup truck as a reference given its common features are more easily discernible than other elements in the scene.

Based on the common height metrics, the ground clearance of a pickup truck is presumed to not be standardized in the prompt. However, if we approximate the ground clearance of this type of truck to be similar to that of an SUV (about 0.3m), we can observe the following from the image:

- The floodwater reaches a point on the truck's wheels that is above the ground clearance but below the bottom of the door, which is 0.7m high for an SUV.

Given these observations and without the exact metrics for a pickup truck, we can make an educated guess:

- The water level is above the ground clearance of 0.3m but below the door's bottom at 0.7m. In the floodwater's visible relationship to the truck's wheel and considering the proportions of the wheel and tire, it seems that the water is just reaching or just above the halfway mark of the wheel height.

Taking these factors into account and providing an estimation that aligns with the given metrics, I would estimate the floodwater depth to be around 0.6 meters. This is a discrete estimate based on the provided truck car metrics and the visual information available in the image.

**Figure 5. Sample of FloodDepth-GPT response (Appendix 2 provides more samples).**

Figures 6 – 8 present examples where FloodDepth-GPT's estimations diverged from human assessments. The discrepancies can be attributed to variations in the estimation points, insufficient criteria, and incorrect water level identification. Figure 6, where a human served as the reference object, the model assessed the water level to be above the knee (approximately at the mid-thigh), while in reality, the water level on the human appears to be approximately at knee level. Future research could address this issue through enhanced prompt engineering to improve the perspective through which the model observes the water level. Additionally, discrepancies in estimations occurred from the limited criteria used for estimation by both humans and the GPT model. For instance, in Figure 7, street signs were utilized as reference points without specifying their heights in the estimation criteria, leading to variations in the estimations made by humans and the GPT model. Figure 8 demonstrates significant variation in estimations, likely owing to divergent observation points by human observers and the GPT model, such as the center of the road or roadside. This emphasizes the challenge of precisely estimating flood depths due to variations in the terrain and the depth within a flood scenario.

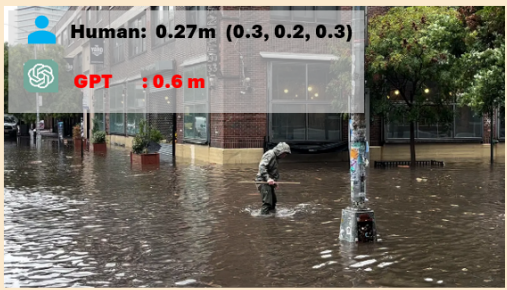

To estimate the floodwater depth in the image, we can use the visible person as a reference. It is difficult to ascertain the person's height precisely, as we do not know if they are male or female, and individuals can vary in height. Nevertheless, given the general height metrics provided, we can make an educated guess.

The water level on the individual appears to be approximately mid-thigh. Assuming an average height between a man and a woman, we could estimate the thigh level to be roughly between the knee and the waist, which gives us a range from 0.4m to somewhere between 0.8m and 0.9m. We can estimate the mid-thigh level to be around 0.6 meters (about the halfway point between the knee height and the low end of the waist height range).

Therefore, based on the information visible in the image and the given human height metrics, I would estimate the floodwater depth to be approximately 0.6 meters.

**Figure 6. Variation in estimations due to wrong identification of water level**

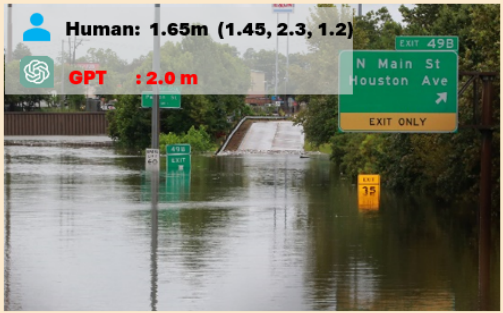

The image shows a flooded roadway with roadway signage partially submerged. To estimate the floodwater depth, I will use the roadway signage as a reference since there are no humans, cars, trucks, SUVs, or buses visible.

Looking at the speed limit sign on the left side of the image and knowing that the total height of a stop sign, including the pole, is approximately 2.9 meters, and the height of the sign itself is 0.9 meters, we can assess the depth of the water based on how much of the pole is submerged.

Assuming the speed limit sign has similar dimensions to the stop sign, considering that the water level is reaching just below the sign plaque (which starts at around 2 meters from the ground if we infer that from the typical heights of a stop sign pole), we can deduce that the water depth is around 2 meters at the sign's location.

Therefore, the estimated floodwater depth in this area, using the submerged sign as a reference, is approximately 2 meters.

**Figure 7. Variation in estimations due to insufficient criteria for estimation**

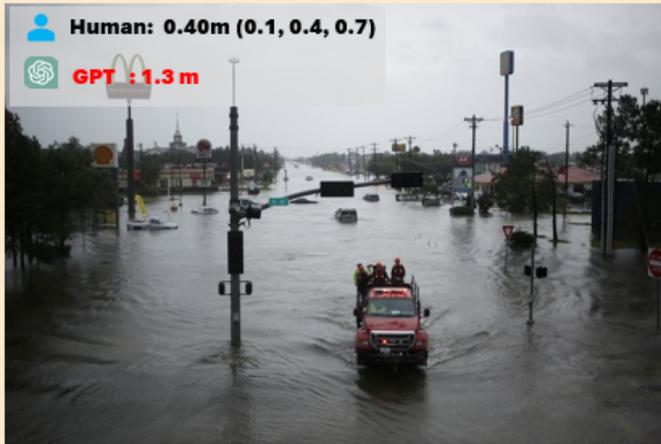

does appear to be above the height from the ground to the top of the hood, which is 1.3 meters. Therefore, it is reasonable to estimate that the floodwater depth around the truck is around or slightly greater than 1.3 meters but less than 1.4 meters.

Farther in the background, there is a sedan car with the roof almost completely submerged. Given that a sedan's overall height from the ground to the roof is 1.4 meters, we can infer that the floodwater is at least that deep where the sedan is located.

Lastly, looking at the street sign on the left side of the photo, the water level reaches about halfway up the pole. Since the total height of the sign is 2.9 meters, and the top of the stop sign plaque is 0.9 meters, the floodwater depth at that point appears to be around the middle of the pole height, which would suggest a depth that could be approximately around 2 meters or slightly less, considering the visible unflooded length of the sign pole above the water.

Taking all these observations into account, and aiming for a discrete estimate, the floodwater appears to range in different depths due to the varying distances and objects. For this particular estimation exercise, considering a central point like the rescue vehicle for a more consistent reference, I would estimate the floodwater depth to be around 1.3 meters where the truck is situated since it is the most reliable reference point in the immediate visible area. This is under the assumption that the truck is similar to a typical truck with the specifications provided. However, it should be noted that this is a rough estimation and the actual depth can vary at different points in the image.

**Figure 8. Variation in the observation point within the flood photo**

## 4    Discussion and Future Research

This study introduces an automated approach to estimating floodwater depth by leveraging the ability of GPT-4. This method utilizes structured prompts to analyze flood photos and estimate flood depth. In contrast to conventional computer vision and deep learning methods that depend on specific pre-trained objects, FloodDepth-GPT can automatically identify water levels based on reference objects on flood photos. This approach streamlines the estimation process and enhances the rapidity of flood inundation mapping.

Results from this study reveal that the proposed method can utilize a variety of common reference objects in flooding photos. This enhances the method's versatility and makes this approach applicable in different flood scenarios. To the best of our knowledge, previous research has largely focused on models trained to recognize specific objects within flood photos, such as humans, stop signs, vehicles, etc. For example, J. Li et al., (2023) developed an object detection model that could only identify humans in flood photos used in their analyses. Analogously, some other studies only utilized photos containing stop signs (Alizadeh and Behzadan, 2023; Song and Tuo, 2021a). Although previous methods showed promising results, these techniques can only be applied to photos containing objects on which their models were trained, thereby restricting the utility of such models on a broader scale.

We think the findings in this study are transformative. The GPT model's ability to interpret different urban and natural elements in images opens new possibilities for automatic environmental asessmennts. Such detailed assessments are crucial for urban planning, disaster preparedness, and climate change studies can be achieved through AI-driven analysis. Moreover, this study highlights FloodDepth-GPT as an example of Explainable AI, which provides the reasoning behind the model's decision-making. This illustrates the transparency of the method, which is crucial for fostering trust and understanding in AI-driven environmental assessments.

The results of this study indicate that the proposed method is reliable in its estimations, demonstrating a mean average error within an fair range for estimating floodwater depths. However, further examinations revealed the presence of outliers in the estimations. While some outliers do not necessarily indicate errors by the GPT model but rather reflect differences in the observation points of estimation within the photo (see Figure 6 – 8), further refinements in future studies could include the introduction of more detailed criteria for estimation. Another limitation of the proposed approach is the differences in the heights and dimensions and designs of reference objects across different regions. This can limit the model's ability to provide uniform estimations across diverse geographic locations. Future studies can focus on calibrating the model to account for regional variations in reference objects.

One potential future research avenue is photo localization, as the floodwater depth is only useful when accurate geolocation of the photo is known, at least at street level. If flood photos and their geolocations can be obtained in real-time from various sources, the proposed approach could provide water depth estimates promptly, which opens the possibility of producing flood inundation maps in real-time. The most straightforward way of photo localization is to extract location information from the photo if such information is available in the metadata or from a geotagged post associated with the photo (Ning et al., 2020; Huang et al., 2019, 2020). For example, some social media platforms allow users to geotag photo location using the phone's built-in Global Positioning System (GPS) or manually select the street name and address. A big challenge, though, is to locate the photo based on photo content only, where the location can be obtained by retrieving similar photos in a large database with localized photos such as street view images (e.g., Zhang et al., 2020). It is particularly challenging to locate flood photos as such photos often have inundated features, resulting in potential mismatching to the existing photos. Thus, innovative methods are required to match the flood and non-flood photos, such as the semantic scene graph (Yoon et al., 2021).

## 5  Conclusion

Floodwater depth estimation plays a pivotal role in the effective management of floods, facilitating informed disaster response and strategic planning. Utilizing the advanced capabilities of large pre-trained multimodal models (GPT-4 in this study), this paper introduces a novel method for automatically determining floodwater depths from photographs related to flooding. The GPT-4 model was customized to estimate water depths using recognized reference objects in the photo by providing specific instructions related to this task.

The findings of this study indicate that the proposed method holds considerable promise for estimating floodwater depths from photographs. In comparison to prior research in this domain, our study demonstrates a universal pipeline for floodwater depth estimation rather than training various individual models on different reference objects. This versatility results in a more efficient and cost-effective process. Additionally, the method can significantly enhance the speed of flood mapping. Such information gives decision-makers and the community at risk a rapid situation awareness of the extent and impact of flooding, which is essential for effective disaster management and decision-making. However, the study also acknowledges minor discrepancies between the model's estimations and those derived by humans. Future research could enhance this methodology through enhanced prompt engineering and the introduction of additional criteria for more accurate estimations.

As we navigate the future of flood management, AI-driven insights become paramount. This approach, leveraging the power of AI and computer vision, emerges as a means for shaping resilient communities. It not only equips emergency responders and planners with the tools needed for rapid, data-driven decision-making but also fosters the art of innovation in disaster management.

**Appendix 1. Prompt used in designing FloodDepthGPT**

- Flood related photos will be an input. Estimate the floodwater depth based on visible reference points in this image. In estimating the flood water depth, consider the following height metrics for common features:

- For human, consider these height metrics: Men Total height = 1.75m, Knee height = 0.4m, Waist height = 0.9m, Shoulder height = 1.4m; Women: Total height = 1.60m, Knee height = 0.4m, waist height = 0.8m, Shoulder height = 1.4m.

- For sedan cars, consider these height metrics: the overall height from the ground to the roof is 1.4m, the ground clearance is approximately 0.2m, height from ground to the bottom of the door is 0.6m, height from the ground to the top of the hood is 1.0m, and the height from the ground to the bottom of the window is 0.8m.

- For a truck car, consider these height metrics: the overall height from the ground to the roof is 1.8 meters, the ground clearance is approximately 0.5 meters, the height from the ground to the bottom of the door is 0.8 meters, the height from the ground to the top of the hood is 1.3 meters, the height from the ground to the bottom of the window is 1.4 meters.

- For a SUV car, consider these height metrics: the overall height from the ground to the roof is 1.7 meters, the ground clearance is about 0.3 meters, the height from the ground to the bottom of the door is 0.7 meters, the height from the ground to the top of the hood is 1.0 meter.

- For a bus, consider these height metrics: the overall height from the ground to the roof is 3.2 meters, the ground clearance is approximately 0.7 meters, the height from the ground to the bottom of the door is 1.0 meter, the height from the ground to the bottom of the window is 2.0 meters.

- For street signage (including stop signs), the dimension of a stop sign (the sign is a red octagon) is 0.9m by 0.9m, while the vertical measurement from ground to the top of the stop sign, indicating the total height of the stop sign plaque including the pole is 2.9m. Avoid the reflection of the stop sign in the water. Also, use any other features as a secondary reference.

- Based on the water height against the different parts of each features, and the average height metrics, estimate the depth of the water.

- Provide the estimated floodwater depth in meters.

- Give estimation as a discrete number and not interval.

**Appendix 2: More sample responses from FloodDepth-GPT**

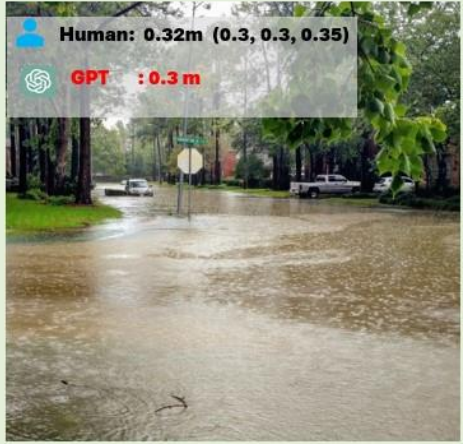

To estimate the floodwater depth, we will use the visible sedan car as a reference point. It appears that the water level is near the top of the car's wheels but does not reach the bottom of the door. According to the height metrics provided for a sedan car, the ground to the bottom of the door is 0.6 meters. Since the water has not reached that point, it is safe to estimate that the water depth is less than 0.6 meters.

However, looking at the wheels, they are typically about 0.6 meters in diameter, and it seems that the water is just at the central hub or a bit higher, indicating that the water level might be around the halfway mark of the wheel's height, which would be approximately 0.3 meters from the ground.

Given this and the lack of details that might suggest otherwise, an estimated discrete water depth could be roughly 0.3 meters.

*Photo source: https://vaccineimpact.com/wp-content/uploads/sites/5/2017/08/Hurricane-Harvey-flooded-street.jpg*

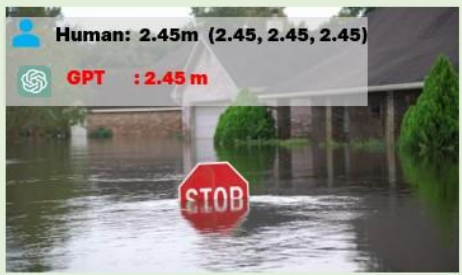

To estimate the depth of the floodwater in this image, we can use the visible stop sign as a reference point. Given that the total height of a stop sign, including the pole, is typically 2.9 meters, and it seems that only the upper part of the sign is visible above the water, we can make an estimation.

From the image, it appears that about half of the stop sign is submerged, with the bottom of the stop sign (0.9m by 0.9m) being below the water level. If we consider that around half of the sign's octagonal face and some portion of the pole beneath it are underwater, we can estimate the water depth as follows:

- The exposed part of the sign is approximately half of the sign's face, which is about 0.45 meters, plus a visible portion of the pole above the sign.

- Knowing the total height of the stop sign with the pole is 2.9 meters, if about 0.45 meters of the sign is exposed above the water, then the submerged portion of the pole and sign can be estimated as 2.9 - 0.45 = 2.45 meters.

This suggests an approximate water depth of about 2.45 meters. It is important to note that this estimation assumes the stop sign is a standard size and that it's fully upright and not tilting, which could affect the perceived water depth.

*Photo source: https://ak.picdn.net/shutterstock/videos/32103613/thumb/1.jpg*

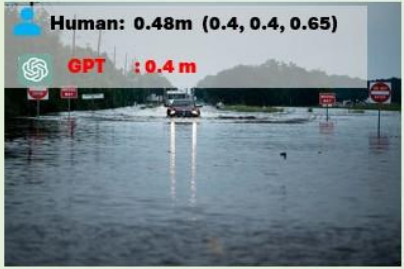 To estimate the floodwater depth in this image, I will use the reference points provided and visible in the photo. Based on the image, we can see a truck moving through the water, roadside signage, and some environmental features.

However, given the size of the truck relative to the known metrics, we can attempt to estimate the water depth. The water level appears to be below the bottom of the truck's door, so it is less than 0.8 meters. Since we cannot see exactly where the water reaches on the truck, I will make an estimation based on the visible water line and the truck's characteristics.

Considering that the truck seems to be a typical one with the metrics provided and that a small section of the wheel and some of the vehicle's body are visible above the water, the depth could be estimated around halfway up the tire or a bit higher. Given the truck's ground clearance (0.5 meters), I would estimate the water depth to be around 0.4 meters deep. However, I cannot confirm the exact model and scale of the truck base against the standard height metrics provided.

The road signage seems fairly far away, and without knowing if they're at a higher elevation, using them for a precise estimate is not reliable.

Given the available information, the estimated floodwater depth, considering the metrics provided and observed water levels, is approximately 0.4 meters.

*Photo source: https://d.newsweek.com/en/full/657216/8-30-17-hurricane-harvey-flooding.jpg*

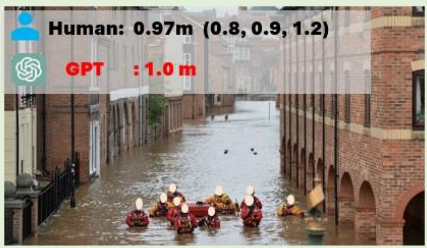 To estimate the floodwater depth in the image, I'll use the reference points provided in the image and compare them against the height metrics for humans and common features.

Starting with the group of rescuers in the water, the floodwater seems to reach approximately between the waist and chest level of the standing individuals. Given that these could be a mix of men and women:

- If it is at waist level for an average man (0.9 meters), the depth would be slightly less because the water does not reach their shoulders.

- If it is at waist level for an average woman (0.8 meters), the depth would be slightly more because the water seems closer to chest level.

The visible heights of the individuals vary, however the water level is clearly above waist height for all visible individuals. Therefore, estimating conservatively and considering the group's postures and the visible water line on their gear:

Estimated floodwater depth: 1 meter

It is important to note that due to the perspective and lack of clear reference objects (like the vehicles or signage described), this is an approximation. Actual water depth can vary depending on the exact heights of the individuals and the angle of the photograph.

*Photo source: https://ichef.bbci.co.uk/news/976/cpsprodpb/4F47/production/_87559202_hi030779656.jpg.webp*

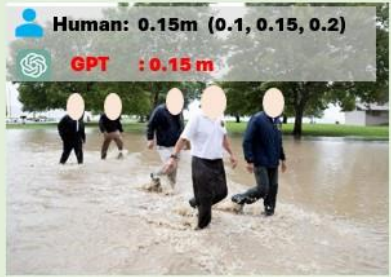

To estimate the floodwater depth from the image provided, I can use the height metrics for an average man, as the image shows a group of men wading through floodwater. The water level appears to be below their knee height but above their ankles.

Considering the average knee height for a man is about 0.4 meters, and the water does not reach that high, we need to estimate a value between the ankle and just below the knee. Given that the lower leg (from the ankle to the knee) would typically comprise slightly less than half the total length of the leg (from the hip joint to the ankle), I'll estimate the water level to be closer to one-third of the way up the lower leg.

Therefore, if we estimate that one-third of the distance from the ankle to the knee is about 0.13 meters (one-third of 0.4 meters), we can add this to the average ground clearance of a man's shoe, which may be about 0.02 meters, giving us an approximate floodwater depth of 0.15 meters (15 centimeters).

Estimated floodwater depth: 0.15 meters.

*Photo source: https://images.hamodia.com/hamod-uploads/2018/08/14170042/30165176018_f517fbfee7_o-1024x727.jpg*

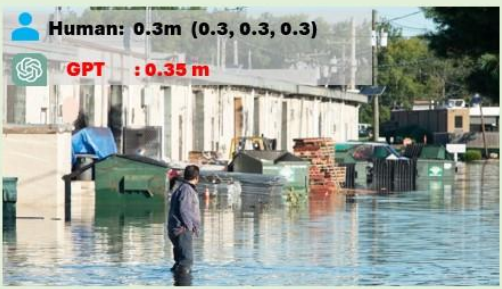

In the provided image, there is a person standing in the floodwaters. Based on the information you've given regarding average human heights, we can estimate the floodwater depth with reference to the individual in the photo.

Judging from the water level in relation to the person's body, it appears that the water is just below the knee height. Since we don't know the exact height of this individual and can't assume they are exactly average, we'll use the given average metrics for knee height.

For men, the knee height is indicated to be 0.4m and for women, it is also 0.4m. The floodwater level appears to be just below this point, so the estimated depth would be somewhat less than the knee height.

Given this observation, an estimated floodwater depth in this photo could be approximately 0.35 meters, as it is slightly below the average knee height.

*Photo source: https://www.usatoday.com/gcdn/presto/2021/09/03/PNJM/a61d84b8-e6f6-4a6e-a712-de6e9fd3627c-090321_Fairfield_WeatherTZ_1816.JPG*

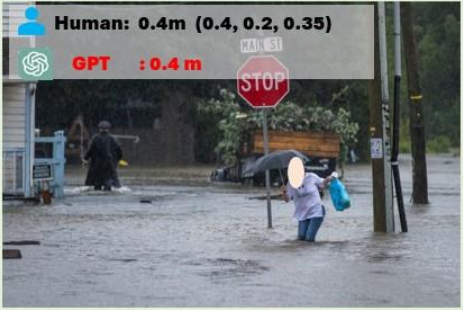
 In the image provided, you can see a person walking through the floodwater, and there is also a stop sign visible. The person appears to be an adult, and the water level reaches approximately to their knee area. Taking into account the average knee height for an adult (as provided for both men and women is 0.4m), this gives us a rough estimate of the floodwater depth.

As for the stop sign, the bottom of the stop sign plaque is visibly above the water level by a considerable distance, making it a less useful reference point for determining water depth in this scenario.

Based on the information given and the visible reference of the floodwater reaching to the knee height of the person, an estimated floodwater depth can be given as approximately 0.4 meters.

*Photo source: https://www.ydr.com/gcdn/presto/2021/09/02/PPYR/318ffd32-0fe7-4d4e-8945-41c6103550a9-HES-HJ-090121-McSherrystown-Flooding-02.jpg?width=660&height=441&fit=crop&format=pjpg&auto=webp*

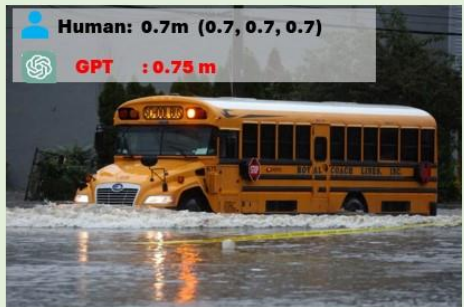
 To estimate the floodwater depth in the image, I ll use the metrics provided for a bus.

- The ground clearance of a bus is approximately 0.7 meters.
- The height from the ground to the bottom of the door is 1.0 meter.

In the image, it appears that the water level is above the ground clearance of the bus but has not reached the bottom of the door. The floodwater depth can, therefore, be estimated to be between 0.7 meters and 1.0 meters.

To provide a specific number rather than a range, since the water is not up to the door level, I'll estimate the floodwater depth to be closer to the ground clearance of the bus. A precise estimation might be challenging as the exact point on the bus where the water ends is not clearly visible, but if I assume the water is just above the ground clearance mark, I would estimate the floodwater depth to be around 0.75 meters.

*Photo source: https://www.theguardian.com/us-news/2023/sep/30/fema-government-shutdown-weather-disasters#img-1*

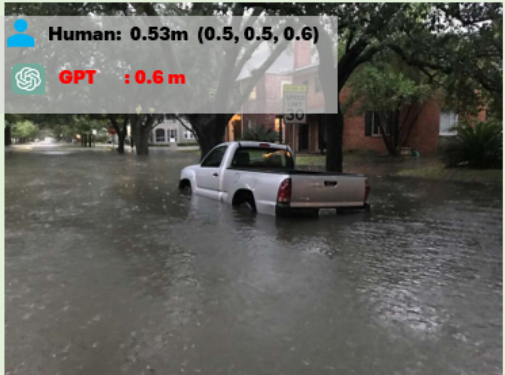 To estimate the floodwater depth in the provided image, let's use the pickup truck as a reference given its common features are more easily discernible than other elements in the scene.

Based on the common height metrics, the ground clearance of a pickup truck is presumed to not be standardized in the prompt. However, if we approximate the ground clearance of this type of truck to be similar to that of an SUV (about 0.3m), we can observe the following from the image:

- The floodwater reaches a point on the truck's wheels that is above the ground clearance but below the bottom of the door, which is 0.7m high for an SUV.

Given these observations and without the exact metrics for a pickup truck, we can make an educated guess:

- The water level is above the ground clearance of 0.3m but below the door's bottom at 0.7m. In the floodwater's visible relationship to the truck's wheel and considering the proportions of the wheel and tire, it seems that the water is just reaching or just above the halfway mark of the wheel height.

Taking these factors into account and providing an estimation that aligns with the given metrics, I would estimate the floodwater depth to be around 0.6 meters. This is a discrete estimate based on the provided truck car metrics and the visual information available in the image.

*Photo source: https://s.hdnux.com/photos/65/15/12/13947547/6/920x920.jpg*

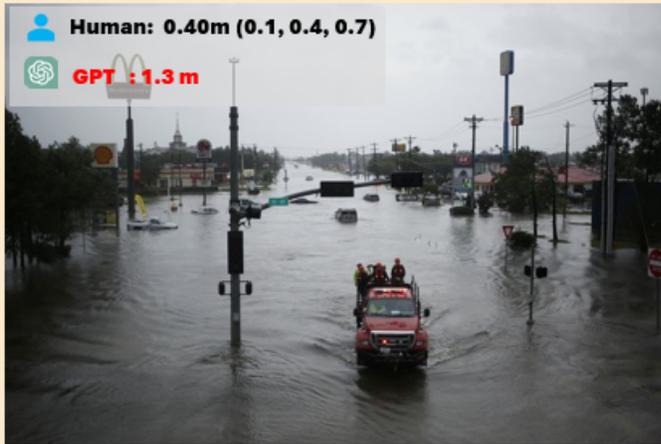
To estimate the floodwater depth in this image, I will consider the visible reference points and the height metrics provided.

In the foreground, there is a rescue vehicle that appears to be a truck. Observing the water level in relation to the truck, the water is significantly below the bottom of the vehicle's window, which is stated to be at a height of 1.4 meters for a truck. However, the water does appear to be above the height from the ground to the top of the hood, which is 1.3 meters. Therefore, it is reasonable to estimate that the floodwater depth around the truck is around or slightly greater than 1.3 meters but less than 1.4 meters.

Farther in the background, there is a sedan car with the roof almost completely submerged. Given that a sedan's overall height from the ground to the roof is 1.4 meters, we can infer that the floodwater is at least that deep where the sedan is located.

Lastly, looking at the street sign on the left side of the photo, the water level reaches about halfway up the pole. Since the total height of the sign is 2.9 meters, and the top of the stop sign plaque is 0.9 meters, the floodwater depth at that point appears to be around the middle of the pole height, which would suggest a depth that could be approximately around 2 meters or slightly less, considering the visible unflooded length of the sign pole above the water.

Taking all these observations into account, and aiming for a discrete estimate, the floodwater appears to range in different depths due to the varying distances and objects. For this particular estimation exercise, considering a central point like the rescue vehicle for a more consistent reference, I would estimate the floodwater depth to be around 1.3 meters where the truck is situated since it is the most reliable reference point in the immediate visible area. This is under the assumption that the truck is similar to a typical truck with the specifications provided. However, it should be noted that this is a rough estimation and the actual depth can vary at different points in the image.

*Photo source: https://pyxis.nymag.com/v1/imgs/40f/dae/449daf31e4753f3f74733fcfc9b082efa5-30-harvey-5.rhorizontal.w700.jpg*

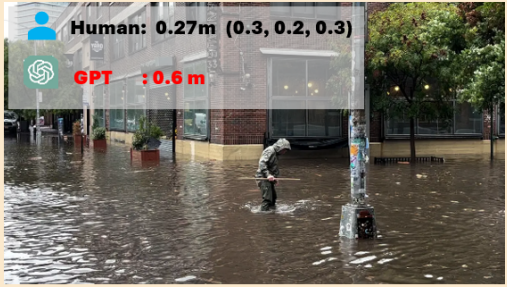
To estimate the floodwater depth in the image, we can use the visible person as a reference. It is difficult to ascertain the person's height precisely, as we do not know if they are male or female, and individuals can vary in height. Nevertheless, given the general height metrics provided, we can make an educated guess.

The water level on the individual appears to be approximately mid-thigh. Assuming an average height between a man and a woman, we could estimate the thigh level to be roughly between the knee and the waist, which gives us a range from 0.4m to somewhere between 0.8m and 0.9m. We can estimate the mid-thigh level to be around 0.6 meters (about the halfway point between the knee height and the low end of the waist height range).

Therefore, based on the information visible in the image and the given human height metrics, I would estimate the floodwater depth to be approximately 0.6 meters.

*Photo source: https://media.zenfs.com/en/the_independent_577/a55c48cbe6dd3449181b2e412b6707c6*

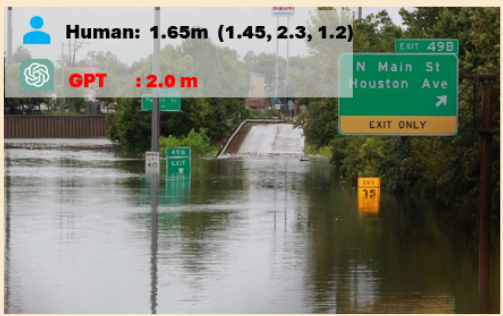
The image shows a flooded roadway with roadway signage partially submerged. To estimate the floodwater depth, I will use the roadway signage as a reference since there are no humans, cars, trucks, SUVs, or buses visible.

Looking at the speed limit sign on the left side of the image and knowing that the total height of a stop sign, including the pole, is approximately 2.9 meters, and the height of the sign itself is 0.9 meters, we can assess the depth of the water based on how much of the pole is submerged.

Assuming the speed limit sign has similar dimensions to the stop sign, considering that the water level is reaching just below the sign plaque (which starts at around 2 meters from the ground if we infer that from the typical heights of a stop sign pole), we can deduce that the water depth is around 2 meters at the sign's location.

Therefore, the estimated floodwater depth in this area, using the submerged sign as a reference, is approximately 2 meters.

*Photo source: https://cdn.theatlantic.com/thumbor/7VUUiVJMgYrI19Rbl_1ltA4SlME=/0x146:3600x2171/976x549/media/img/mt/2017/08/RTX3DKUO/original.jpg*